\documentclass[11pt]{article}

\usepackage[preprint]{acl}

\usepackage{times}
\usepackage{latexsym}

\usepackage[T1]{fontenc}

\usepackage[utf8]{inputenc}

\usepackage{microtype}

\usepackage{inconsolata}

\usepackage{graphicx}
\usepackage{amsmath}
\usepackage{amssymb}
\usepackage{algorithm}
\usepackage{algorithmic}
\usepackage{booktabs}
\usepackage{multirow}

\newcommand{\method}{TAPS}
\newcommand{\qcond}{q_{\mathrm{cond}}}
\newcommand{\qreach}{q_{\mathrm{reach}}}

\title{\textit{TAPS:} Target-Aware Prefix Tree Selection for Diffusion-Drafted Speculative Decoding}

\author{
Zhuoyu Wang$^{*}$ \quad
Junnan Huang$^{*}$ \quad
Xinyu Chen$^{\dagger}$ \\
The Hong Kong University of Science and Technology (Guangzhou) \\
\small $^{*}$Co-first authors. \quad
$^{\dagger}$Corresponding author.
}

\begin{document}
\maketitle
\begin{abstract}
Using a diffusion model for parallel drafting is a promising approach for speculative decoding. By predicting tokens at multiple future positions in a single forward pass, diffusion drafters substantially reduce drafting latency. However, this shifts the bottleneck to verification: verifying a single sequence limits acceptance length, while verifying large draft trees incurs excessive target-model latency.
We identify a key mismatch in existing draft-tree methods: existing diffusion-tree methods rank nodes by the marginal probability, ignoring that verification is prefix-conditioned. As a result, they may verify unreachable descendants of rejected prefixes, increasing latency with limited acceptance gains. To address this, we propose \textbf{TAPS}, a target-aware prefix selection method that turns diffusion marginals into path-conditioned acceptance estimates. \method{} then selects a compact prefix-closed subtree under a fixed verification budget, improving the acceptance-cost tradeoff rather than simply expanding the draft tree.
Experiments across diverse datasets and model families demonstrate that TAPS achieves up to 7.9x lossless end-to-end speedup over vanilla autoregressive decoding, outperforming state-of-the-art DFlash and DDTree by 1.36× and 1.74× respectively. Our work is available at https://anonymous.4open.science/r/TAPS-EMNLP2026-53DD
\end{abstract}

\begin{figure}[!t]
    \centering
    \includegraphics[width=0.98\linewidth]{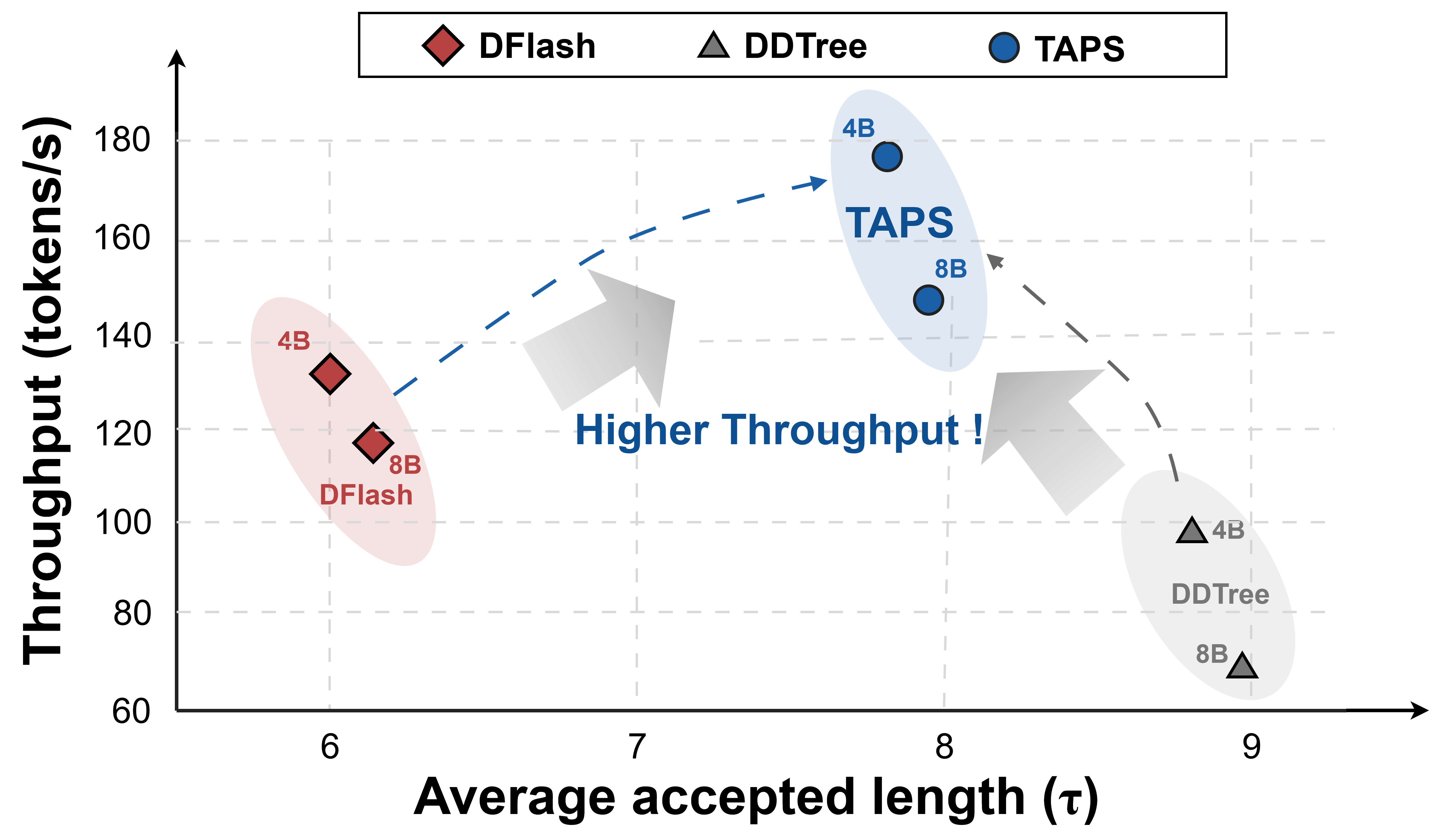}
    \caption{\textbf{Overall throughput--acceptance trade-off.}
    We compare DFlash, DDTree, and \method{} under Qwen3-4B and Qwen3-8B settings, averaged across all benchmarks on A40 GPU.
    \method{} achieves a better throughput--acceptance tradeoff than prior methods, improving throughput while maintaining competitive accepted length.}
    \label{fig:teaser}
\end{figure}

\begin{figure*}[t]
    \centering
    \includegraphics[width=\textwidth]{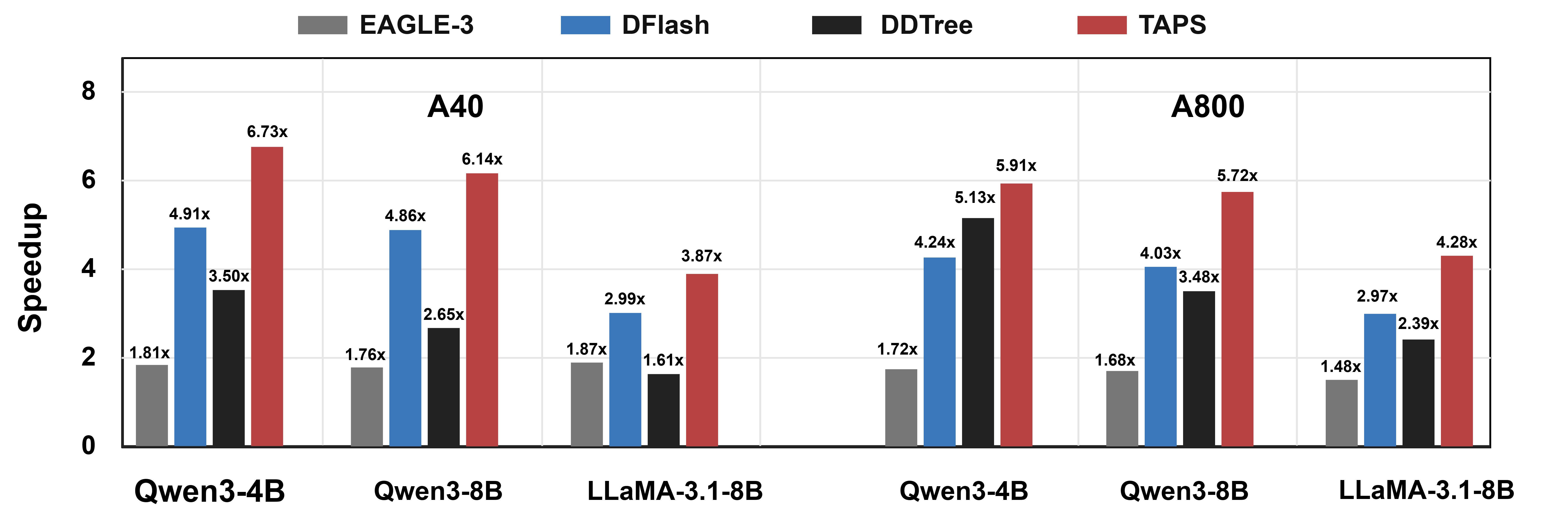}
    \caption{Average speedup across models and GPU platforms. TAPS consistently outperforms EAGLE-3, DFlash, and DDTree across target models and hardware platforms.}
    \label{fig:avg}
\end{figure*}

\begin{figure*}[t]
    \centering
    \includegraphics[width=\textwidth]{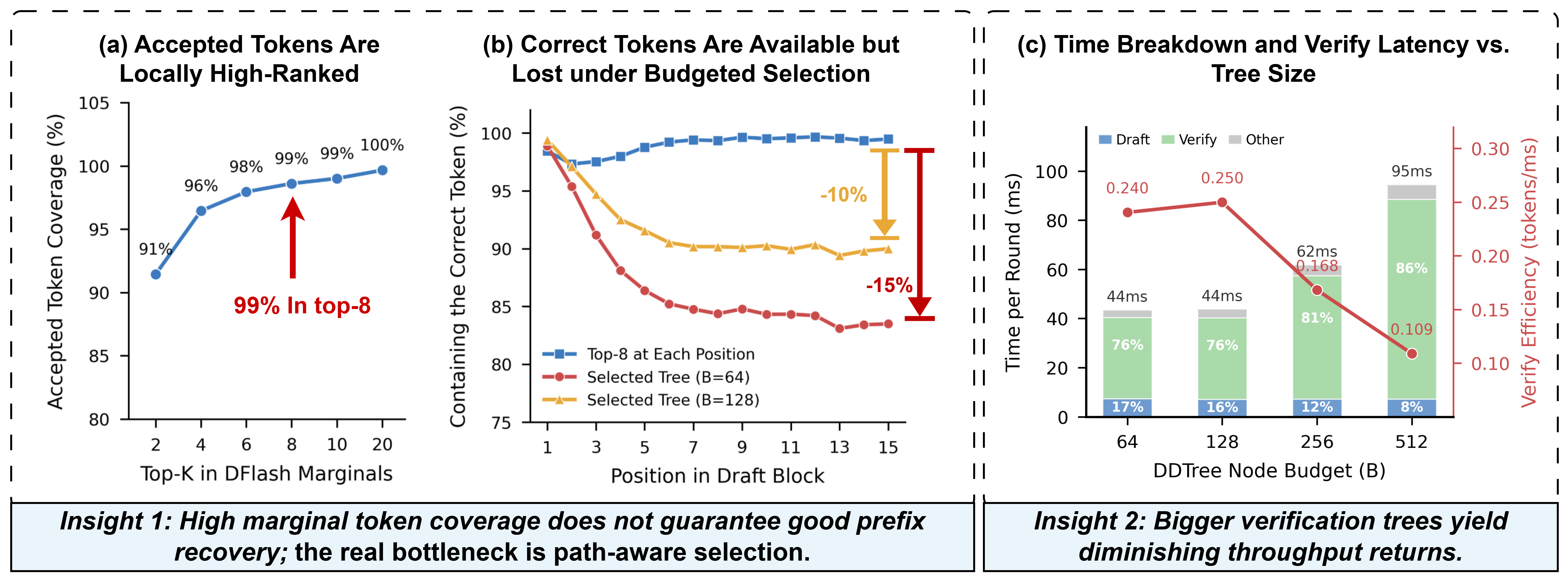}
    \caption{\textbf{Insight 1}: (a) The rank distribution of target-accepted tokens in DFlash marginals. (b) The probability of containing the correct token at each draft-block position under per-position Top-8 selection and selected trees with different budgets. \textbf{Insight 2}: (c) Per-round time breakdown and verification efficiency as the tree node budget increases.All measurements are collected with Qwen3-4B across diverse datasets.}
    \label{fig:motivation}
\end{figure*}

\begin{figure}[!b]
    \centering
    \includegraphics[width=0.98\linewidth]{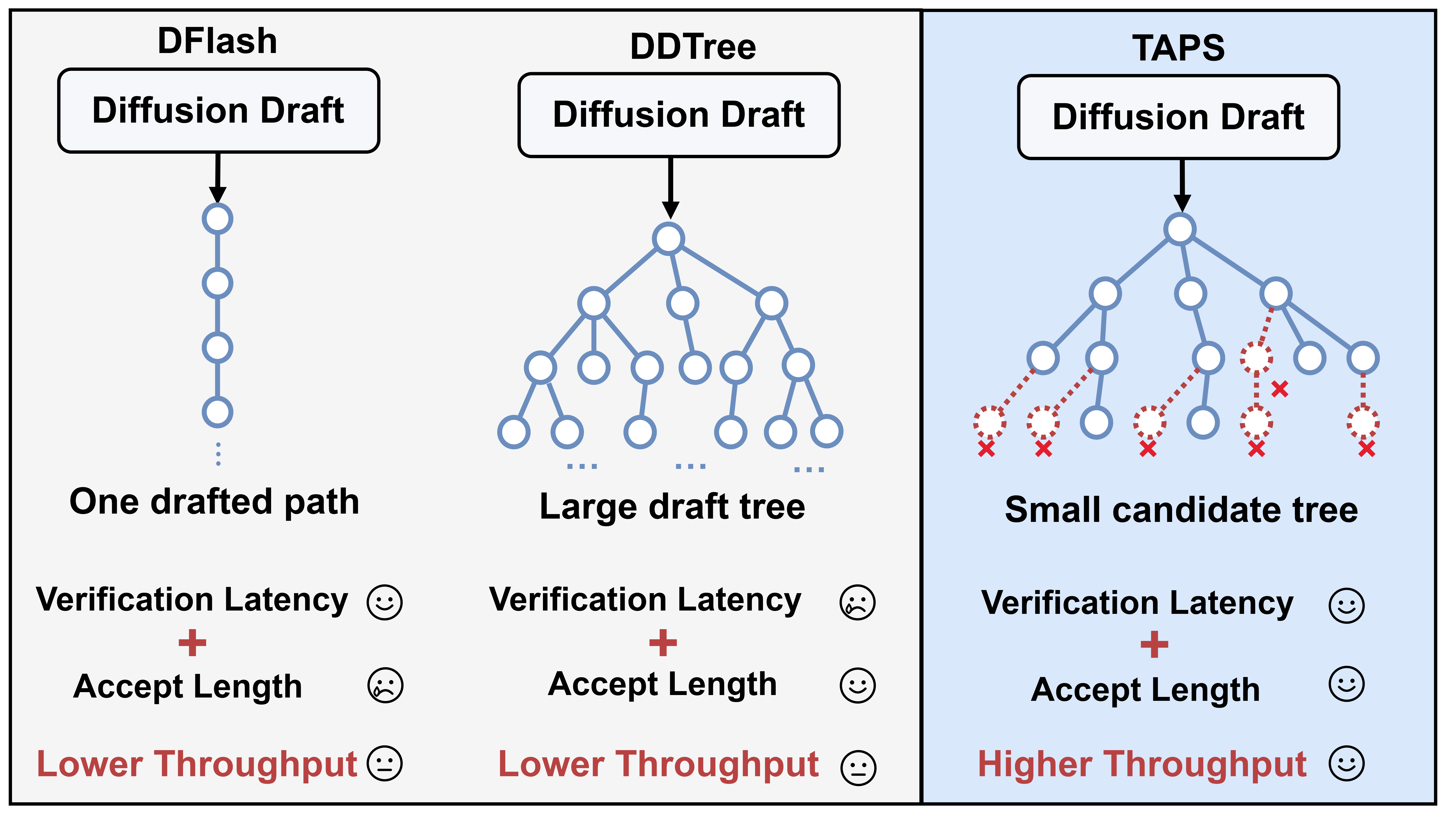}
    \caption{Comparison of DFlash, DDTree, and TAPS.}
    \label{fig:Tree}
\end{figure}

\begin{figure*}[t]
    \centering
    \includegraphics[width=\textwidth]{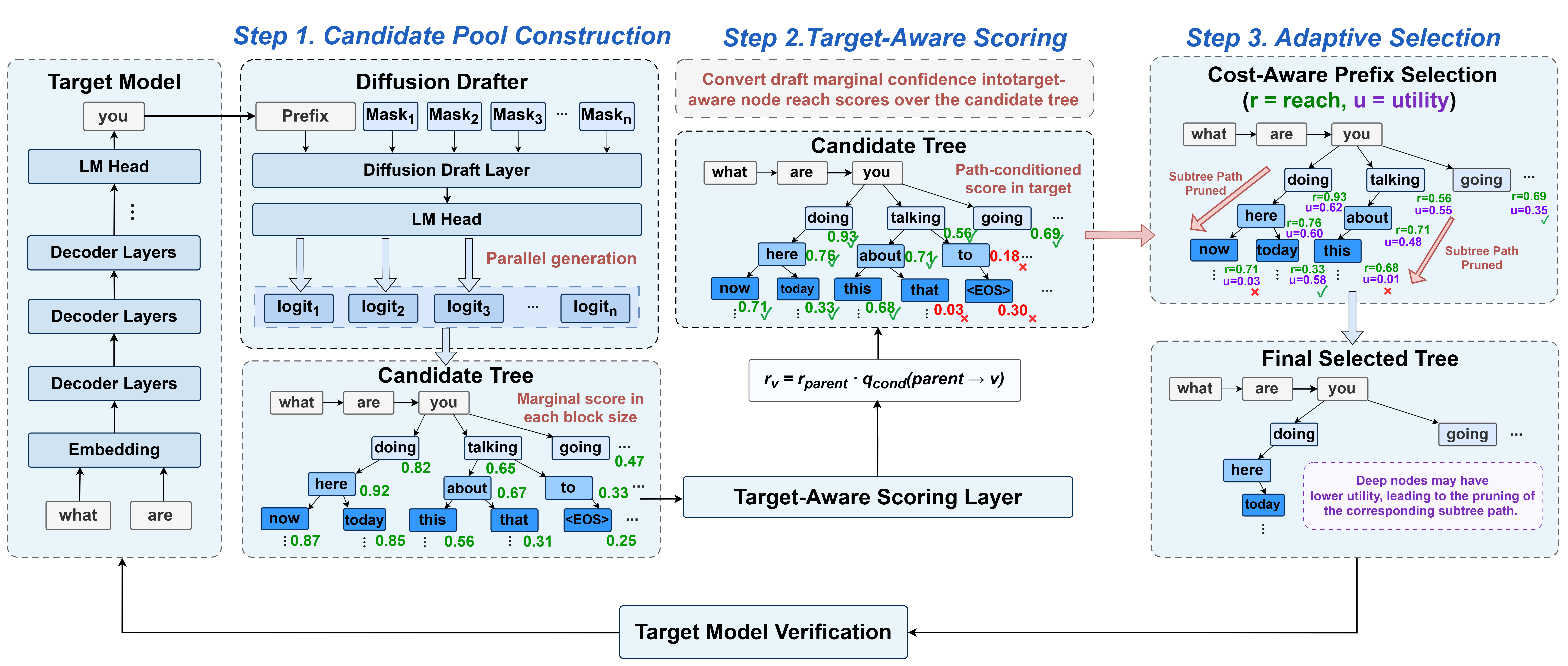}
    \caption{Overview of \method{}. The diffusion drafter first produces a large marginal candidate pool; the target-aware scoring layer converts marginal token confidence into path-conditioned reach probabilities; the adaptive selector then keeps a compact prefix-closed subtree for efficient target-model verification.}
    \label{fig:overview}
\end{figure*}

\section{Introduction}

Speculative decoding has become one of the most effective approaches for accelerating large language model (LLM) inference without compromising output quality~\citep{leviathan2023fast, chen2023accelerating}.
The key idea is to use a lightweight draft model to quickly propose multiple candidate tokens, which the target model then verifies in parallel through a single forward pass.
By amortizing the cost of target-model inference over multiple accepted tokens, speculative decoding achieves significant wall-clock speedup while provably preserving the target model's output distribution.

Diffusion-based drafting has emerged as a particularly promising strategy for speculative decoding.
Unlike autoregressive drafters that must generate candidate tokens one by one, diffusion drafters such as DFlash~\citep{chen2026dflash} predict an entire block of future tokens in a single forward pass, substantially reducing draft latency.
To further improve acceptance length, recent methods organize candidates into tree structures that allow the target model to verify multiple alternative paths simultaneously~\citep{miao2024specinfer, li2024eagle}.
Combined with tree-structured verification, methods like DDTree~\citep{ringel2026ddtree} construct large candidate trees from diffusion draft predictions and verify all branches in parallel, improving average accepted length.

However, diffusion drafting changes the bottleneck: once draft generation becomes cheap, target-model verification dominates end-to-end latency. Tree verification can increase accepted length by checking multiple continuations in a single target-model forward pass, but its cost grows with the number of selected nodes. Since each decoding round commits to only one prefix path, nodes under rejected prefixes add verification latency without contributing accepted tokens. 

 To address this problem, we propose \method{} (\textbf{T}arget-\textbf{A}ware \textbf{P}refix \textbf{S}election), a method that learns to predict which candidate nodes in a diffusion draft tree are most likely to be accepted by the target model, and dynamically selects a compact subtree that improves throughput under a verification budget constraint.
Our main contributions are as follows:

\begin{itemize}
    \item \textbf{Target-aware acceptance modeling.}
    We introduce a lightweight learned scorer that estimates the conditional acceptance probability of each candidate token along its drafted path, replacing draft-confidence-based selection with target-aware scoring.

    \item \textbf{Cost-aware dynamic tree pruning.}
    \method{} propagates acceptance probabilities through the candidate tree and greedily selects a compact prefix-closed subtree that improves expected accepted tokens under a verification budget constraint.

    \item \textbf{Significant end-to-end acceleration.}
    As shown in Figure~\ref{fig:teaser}, \method{} achieves the highest end-to-end throughput while maintaining relatively high average accepted length $\tau$. Figure~\ref{fig:avg} shows that \method{} achieves an average speedup of 5.44$\times$ across three target models and two GPU platforms, improving over DFlash and DDTree by 1.36$\times$ and 1.74$\times$, respectively.
\end{itemize}

\section{Related Work}

\paragraph{Speculative Decoding and Draft Trees.}
Speculative decoding accelerates LLM inference by using draft candidates that are verified by the target model while preserving the target distribution~\citep{leviathan2023fast, chen2023accelerating}.
Existing methods improve candidate generation from different angles, including auxiliary draft models, self-speculative heads, retrieval-based drafting, recurrent drafting, and multi-head feature predictors~\citep{cai2024medusa, ankner2024hydra, cheng2024redrafter, he2024rest, li2024eagle}.
To increase accepted length per verification pass, tree-based methods verify multiple candidate continuations with tree attention, with representative systems including SpecInfer~\citep{miao2024specinfer}, Sequoia~\citep{chen2024sequoia}, and EAGLE-2~\citep{li2024eagle2}.
These works demonstrate that expanding the candidate structure can improve acceptance length, but larger trees also increase target-model verification cost.

\paragraph{Diffusion-Based Drafting.}
Diffusion language models provide a non-autoregressive alternative to left-to-right generation and have been studied in both discrete and continuous text generation settings~\citep{austin2021d3pm, li2022diffusionlm, nie2025llada}.
Recent speculative-decoding methods exploit this parallelism to draft multiple future tokens in one pass.
DFlash~\citep{chen2026dflash} predicts block-level marginal logits, while DDTree~\citep{ringel2026ddtree} constructs large verification trees from these logits.
Other diffusion-style decoding methods explore lightweight drafting, lookahead filling, and autoregressive verification for faster generation~\citep{liu2026dart, cheng2025deer, xu2025lopa}.
These methods substantially reduce draft-generation latency, making target-model verification an increasingly important component of end-to-end decoding time.

\paragraph{Adaptive Verification Efficiency.}
Another line of work improves speculative decoding by adapting speculation effort according to runtime difficulty.
Representative methods reduce wasted computation through early rejection, adaptive stopping, dynamic lookahead, or entropy-based draft control~\citep{pan2025failfast, zhang2025draft, mamou2024disco, liu2024pearl}.
These approaches show that fixed speculation policies are often suboptimal: easy contexts can benefit from more aggressive drafting, while difficult contexts require more conservative verification.
\section{Motivation}
\label{sec:motivation}

\subsection{Limitations of Diffusion Draft Trees}
\label{sec:limitations}

Diffusion draft models reduce drafting latency by predicting multiple future positions in parallel.
Once drafting becomes cheap, however, the bottleneck shifts to target-model verification.
As shown in Figure~\ref{fig:Tree}, DFlash~\citep{chen2026dflash} verifies a single draft path, keeping verification cost low but limiting accepted length.
DDTree~\citep{ringel2026ddtree} expands the diffusion draft into a candidate tree, allowing the target model to verify multiple branches and accept longer prefixes.
However, existing diffusion draft trees still suffer from two limitations.

\textbf{Limitation 1: Marginal selection is path-unaware.}
DDTree selects candidate nodes mainly by the marginal probabilities produced by the diffusion drafter.
This treats candidate quality as a position-wise property, whereas tree verification is prefix-conditioned: a node is useful only if all of its ancestors are accepted.
Thus, a high-marginal-probability token can still have low verification utility if it lies under an incorrect prefix, becoming unreachable during target-model verification.

Figure~\ref{fig:motivation}(a) shows that the target-accepted token appears within the drafter's top-8 marginal candidates in 99\% of cases, indicating that the local candidate pool is usually sufficient.
However, Figure~\ref{fig:motivation}(b) shows that budgeted prefix-closed trees still lose many correct tokens, especially at later positions, with a 10--15\% drop.
This suggests that correct tokens are often available locally but placed under prefixes that diverge from the target-model accepted path.
Therefore, effective tree selection should evaluate each token together with its prefix, rather than ranking nodes independently by position-wise marginal probability.

\textbf{Limitation 2: Larger verification trees provide diminishing throughput returns.}
Expanding the tree can improve coverage, but each decoding round can ultimately commit only one accepted prefix.
Thus, many verified branches are discarded, while tree-attention and target-model verification cost grow with the number of selected nodes.
Figure~\ref{fig:motivation}(c) shows that increasing the tree budget from 64 to 512 nodes raises verification latency from 33\,ms to 81\,ms, while verification efficiency steadily declines.
This indicates that the throughput-optimal tree should balance expected accepted length against verification cost, rather than simply using the largest budget.

\subsection{Key Insights}
\label{sec:insights}

These limitations motivate two design principles.
First, diffusion draft tree selection should be path-aware: since correct tokens are often already present in the local candidate pool, the selector should preserve candidates whose prefixes are likely to survive target-model verification.
Second, verification should be cost-aware rather than budget-filling: the selector should choose a compact prefix-closed subtree that maximizes expected accepted tokens per unit verification cost.

Guided by these insights, \method{} converts diffusion marginal candidates into path-conditioned reach estimates with a lightweight target-aware scorer, and then selects a compact prefix-closed subtree under a verification budget.
This aligns candidate selection with prefix-conditioned verification and improves the acceptance--cost tradeoff without simply expanding the draft tree.

\section{Method: \method{}}
\label{sec:method}

Building on the aforementioned observations, we introduce \method{}, which estimates how likely each token is to be accepted in its prefix and selects a compact subtree for verification dynamically.

\subsection{Inference Pipeline}
\label{sec:pipeline}

\method{} operates within the standard speculative decoding loop, alternating between drafting and verification.
Figure~\ref{fig:overview} illustrates a single decoding round, which proceeds in three stages.

In Stage~1, the diffusion drafter generates per-position marginal logits for $d$ future positions in a single forward pass; the top-$K$ tokens at each position are assembled into a candidate tree of up to $N_{\text{pool}}$ nodes.
Stage 2 scores the candidate tree with a lightweight target-aware scorer and propagates edge-level acceptance estimates into path-level \emph{reach probability}. (Section~\ref{sec:scorer}).
In Stage~3, \method{} selects a compact subtree under a verification budget.
The resulting subtree adapts to difficulty, large when the drafter is confident, small when it is not, while balancing draft confidence against verification cost. Then, subtree is verified by the target model via tree attention, preserving the exact output distribution.(Section~\ref{sec:selection})

\subsection{Path-Conditional Acceptance Algorithm}
\label{sec:scorer}

As discussed in Section~\ref{sec:motivation}, ranking candidates by marginal draft probability fails to account for the sequential gating of tree verification.
\method{} instead decomposes scoring at the \emph{edge} level: for each parent--child pair $(u \to v)$, we estimate the conditional acceptance probability
\begin{equation}
    \qcond(u \to v) = \Pr(\text{accept } v \mid \text{reach } u),
    \label{eq:qcond}
\end{equation}
which directly mirrors how the target model evaluates each draft token conditioned on the prefix of previously accepted tokens along the same path.

\paragraph{Scorer design.}
A lightweight scorer maps each candidate edge $(u \!\to\! v)$ to a scalar logit $\ell_{u,v}$.
The input consists of the parent and child token, the edge depth, and draft-model statistics such as the child log-probability and the positional entropy at that depth.
For each parent node, we normalize the logits of its candidate children with a sibling-wise softmax, so that the resulting scores represent their relative conditional acceptance probabilities once the parent prefix has been reached.
This design lets the scorer capture target-model preferences under the same prefix, while still using draft-confidence signals without invoking the target model at inference time.

\paragraph{Reach propagation.}
A high local $q_{\mathrm{cond}}$ does not necessarily imply a high-value candidate, since the token is useful only if its entire prefix can survive verification.
We therefore define the reach probability of a node as the product of conditional acceptance probabilities along its root-to-node path:
\begin{equation}
q_{\mathrm{reach}}(v)
=
\prod_{(u,w)\in \mathrm{path}(v_0,v)}
q_{\mathrm{cond}}(u \!\to\! w).
\end{equation}
This path-level score naturally downweights descendants of weak prefixes and helps TAPS prioritize candidate continuations that are likely to remain reachable during target-model verification, rather than isolated high-confidence tokens.

\subsection{Target-Aware Tree Selection}
\label{sec:selection}

As observed in Section~\ref{sec:motivation}, acceptance length saturates beyond a task-dependent tree size while verification cost continues to grow.
\method{} addresses this with a utility-based greedy selection that automatically adapts tree size to input difficulty.

Each node's \emph{utility} balances its expected contribution (reach probability) against verification cost:
\begin{equation}
    \text{utility}(v) = \frac{\qreach(v)}{\lambda_1 + \lambda_2 \cdot \text{depth}(v)},
    \label{eq:utility}
\end{equation}
where $\lambda_1$ models the per-node overhead of tree attention and $\lambda_2$ penalizes deeper nodes whose KV-cache entries are less likely to be reused.
The selector greedily adds nodes in decreasing utility order, including all ancestors of each selected node to maintain prefix closure, tree-attention verification requires a connected subtree rooted at the current position.
Selection terminates when the budget $N_{\max}$ is reached or no remaining node exceeds a minimum utility threshold $\tau$ (Algorithm~\ref{alg:taps}).

On confident rounds the drafter produces many high-reach candidates and the budget is filled; on uncertain rounds few nodes pass the threshold, yielding a smaller tree that avoids wasted verification.
Since \method{} modifies only the selection stage, the output distribution is preserved exactly as in standard speculative decoding.

\begin{algorithm}[t]
\caption{\method{}: one decoding round}
\label{alg:taps}
\begin{algorithmic}[1]
\REQUIRE Candidate tree $\mathcal{T}$; budget $N_{\max}$; threshold $\tau$
\ENSURE Selected prefix-closed subtree $\mathcal{S}$
\FOR{each parent $j$ in $\mathcal{T}$}
    \STATE $\qcond(j \!\to\! e) \leftarrow \text{Softmax}(\text{Scorer}(\mathbf{f}_e))$ for all $e \in \mathcal{C}(j)$
\ENDFOR
\FOR{depth $d = 1$ \TO $D$}
    \STATE $\qreach(v) \leftarrow \qreach(p(v)) \cdot \qcond(p(v) \!\to\! v)$
\ENDFOR
\STATE $\text{utility}(v) \leftarrow \qreach(v) \,/\, (\lambda_1 + \lambda_2 \cdot \text{depth}(v))$
\STATE $\mathcal{S} \leftarrow \{\text{root}\}$
\WHILE{$|\mathcal{S}| < N_{\max}$ and $\exists\, v \notin \mathcal{S}$ with $\text{utility}(v) > \tau$}
    \STATE $v^* \leftarrow \arg\max_{v \notin \mathcal{S}} \text{utility}(v)$
    \STATE $\mathcal{S} \leftarrow \mathcal{S} \cup \{v^*\} \cup \text{ancestors}(v^*)$
\ENDWHILE
\RETURN $\mathcal{S}$
\end{algorithmic}
\end{algorithm}

\subsection{Training the Scorer Layer}
\label{sec:training}

The scorer must predict the target model's acceptance behavior without access to the target model at inference time.
We achieve this through offline distillation of recorded verification outcomes, following the standard idea of transferring teacher-model behavior into a lightweight student predictor~\citep{hinton2015distilling}.


\paragraph{Training objective.}

The scorer is trained to identify high-value candidate paths for target-model verification.
This requires capturing two complementary signals.
First, at each reached prefix, the scorer should model the target model's local preference over candidate children, i.e., which next token is more likely to be accepted under the current prefix.
Second, since verification proceeds along a sequence, the value of a candidate token depends on the path it belongs to.
A locally plausible token may still be unhelpful if it lies on a candidate continuation that the target model is unlikely to follow.
Therefore, the scorer should also assign high reach probability to prefixes that are likely to survive verification as complete candidate paths.
For each recorded trace \(\tau\), we use the target-accepted path as supervision, where \(v_{\tau,d}^{\star}\) denotes the \(d\)-th accepted node, \(v_{\tau,0}\) is the current root, and \(D_\tau\) is the accepted depth.
We train the scorer with the following path-aware distillation objective:

\begin{equation}
\begin{aligned}
\mathcal{L}_{\mathrm{TAPS}}
=&
\sum_{j\in\mathcal{P}}
D_{\mathrm{KL}}\!\left(p_T^j \,\|\, q_\theta^j\right) \\
&+
\lambda
\sum_{v\in\mathcal{T}}
\mathrm{BCE}\!\left(\hat r_v, q_v^{\mathrm{reach}}\right),
\end{aligned}
\label{eq:taps_loss}
\end{equation}

The first term encourages accurate local conditional preference at each reached prefix, while the second term calibrates the accumulated reach probability of the accepted path.
This objective directly aligns the training signal with TAPS's selection criterion: preserving compact prefix-closed subtrees that contain high-value candidate paths.

\section{Experiments}
\label{sec:bibtex}

\begin{figure}[b!]
    \centering
    \includegraphics[width=\columnwidth]{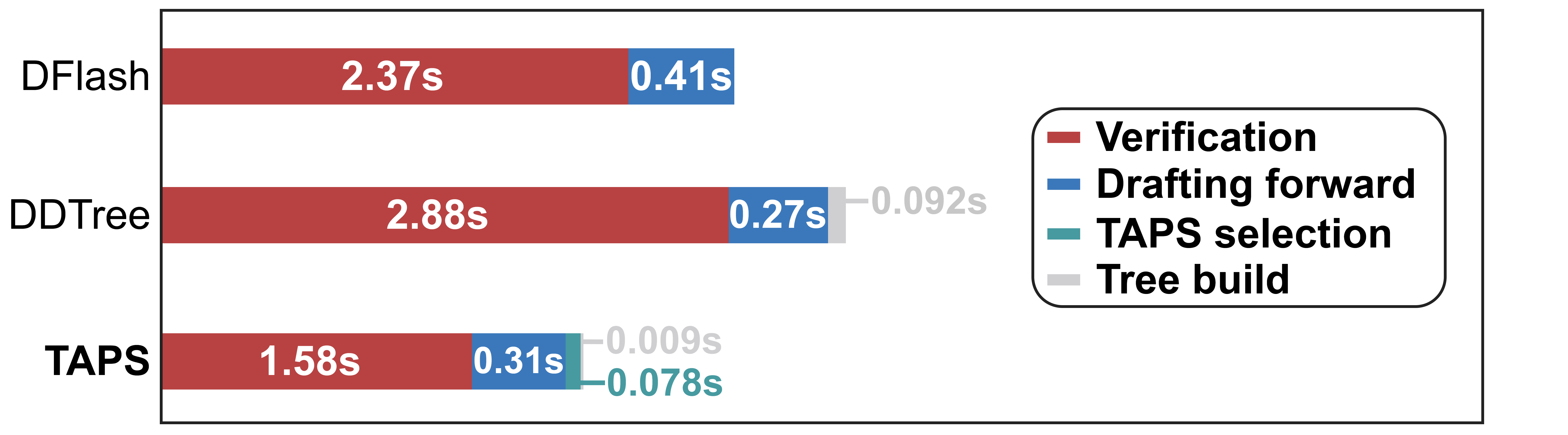}
    \caption{Latency breakdown of DFlash, DDTree, and TAPS. We decompose the total generation time into target-model verification, drafting forward, TAPS selection, and tree construction. }
    \label{fig:Latency}
\end{figure}

\subsection{Experimental Setup}
\paragraph{Models and Datasets.}
We conduct experiments on three mainstream target models, including Qwen3-4B, Qwen3-8B~\citep{yang2025qwen3}, and LLaMA-3.1-8B-Instruct~\citep{grattafiori2024llama}. We evaluate \method{} on seven public datasets in Table~\ref{tab:datasets}. For each model and task, we assess the performance using average acceptance length ($\tau$) and end-to-end decoding speedup over the autoregressive baseline.
\paragraph{Baselines.}
We compare \method{} with autoregressive decoding (baseline), two state-of-the-art diffusion draft speculative decoding methods DFlash~\citep{chen2026dflash} and DDTree~\citep{ringel2026ddtree}. We also include EAGLE-3~\citep{li2025eagle3} as a strong autoregressive draft speculative decoding method to compare \method{}. For a fair comparison, all methods are evaluated with the same prompts, target models, maximum generation length, and decoding configuration. The node budget of DDTree is set to 512. We use the corresponding DFlash draft model and set the draft block size to 16 for Qwen3-4B and Qwen3-8B, and 10 for LLaMA-3.1-8B-Instruct.
 
\begin{table}[t]
    \centering
    \small
    {\renewcommand{\arraystretch}{1.08}
    \begin{tabular*}{0.95\linewidth}{@{\extracolsep{\fill}}p{0.56\linewidth}p{0.44\linewidth}@{}}
        \toprule
        \textbf{Dataset} & \textbf{Task Type} \\
        \midrule
        AIME25~\citep{opencompass2025aime} & Math reasoning \\
        GSM8K~\citep{cobbe2021gsm8k} & Math reasoning \\
        MATH500~\citep{lightman2024lets} & Math reasoning \\
        HumanEval~\citep{chen2021humaneval} & Code generation \\
        LiveCodeBench~\citep{jain2024livecodebench} & Code generation \\
        MBPP~\citep{austin2021mbpp} & Code generation \\
        MT-Bench~\citep{zheng2023mtbench} & Multi-turn dialogue \\
        \bottomrule
    \end{tabular*}
    }
    \caption{Datasets and task types used in the experiments.}
    \label{tab:datasets}
\end{table}

\paragraph{Hardware and Implementations.}
All experiments are conducted  on servers equipped with an NVIDIA A800-SXM4-80GB GPU and an NVIDIA A40-48GB GPU respectively. All methods are evaluated under a single-request decoding setting, with the batch size = 1.

\begin{figure}[b]
    \centering
    \includegraphics[width=\columnwidth]{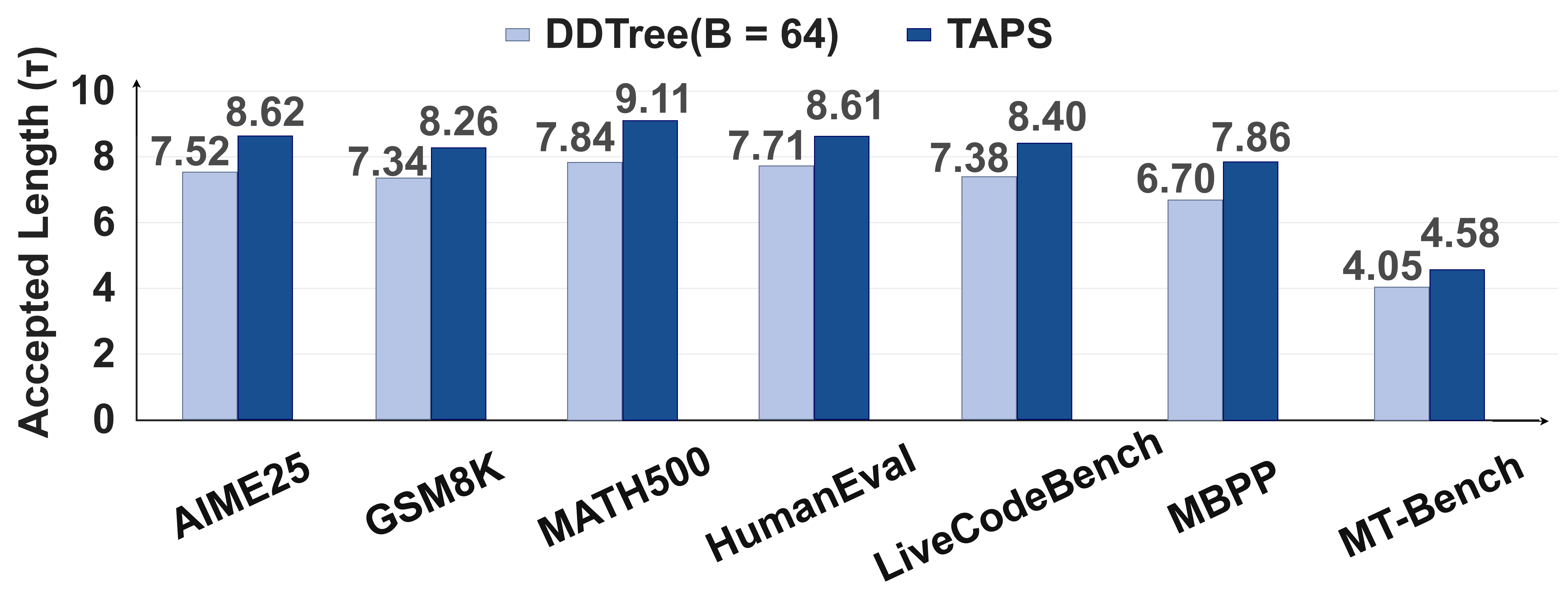}
    \caption{Budget-controlled comparison of acceptance length. We compare DDTree and TAPS under a same budget setting on Qwen3-4B in seven datasets. }
    \label{fig:budget}
\end{figure}

\subsection{Main Results}
\begin{table*}[t]
    \centering
    \setlength{\tabcolsep}{2.5pt}
    {\normalsize
    \renewcommand{\arraystretch}{1.15}
    \resizebox{\textwidth}{!}{
    \begin{tabular}{cc*{8}{cc}}
        \toprule
        \multirow{2}{*}{\raisebox{-0.8ex}{\textbf{Model}}} & \multirow{2}{*}{\raisebox{-0.8ex}{\textbf{Method}}}
        & \multicolumn{6}{c}{\textsc{Math}}
        & \multicolumn{6}{c}{\textsc{Code}}
        & \multicolumn{2}{c}{\textsc{Chat}}
        & \multicolumn{2}{c}{} \\
        \cmidrule(lr){3-8}\cmidrule(lr){9-14}\cmidrule(lr){15-16}\cmidrule(lr){17-18}
        & & \multicolumn{2}{c}{AIME25}
        & \multicolumn{2}{c}{GSM8K}
        & \multicolumn{2}{c}{MATH500}
        & \multicolumn{2}{c}{HumanEval}
        & \multicolumn{2}{c}{LiveCodeBench}
        & \multicolumn{2}{c}{MBPP}
        & \multicolumn{2}{c}{MT-Bench}
        & \multicolumn{2}{c}{\textit{Avg.}} \\
        \cmidrule(lr){3-4}\cmidrule(lr){5-6}\cmidrule(lr){7-8}
        \cmidrule(lr){9-10}\cmidrule(lr){11-12}\cmidrule(lr){13-14}
        \cmidrule(lr){15-16}\cmidrule(lr){17-18}
        \multicolumn{2}{c}{NVIDIA A40-48GB GPU} & Speedup & $\tau$ & Speedup & $\tau$ & Speedup & $\tau$
        & Speedup & $\tau$ & Speedup & $\tau$ & Speedup & $\tau$
        & Speedup & $\tau$ & Speedup & $\tau$ \\
        \midrule
        \multirow{4}{*}{Qwen3-4B} & EAGLE-3
        & 1.79× & 3.05 & 1.99× & 3.30 & 1.83× & 3.08
        & 1.84× & 3.05 & 1.73× & 2.91 & 1.78× & 2.95
        & 1.74× & 3.02 & 1.81× & 3.05 \\
        & DFlash
        & 5.68× & 6.75 & 5.15× & 6.29 & 6.09× & 7.18
        & 5.21× & 6.59 & 5.41× & 6.38 & 4.78× & 5.68
        & 2.85× & 3.15 & 4.91× & 6.00 \\
        & DDTree
        & 3.83× & 9.67 & 3.57× & 9.14 & 4.10× & 10.22
        & 3.58× & 9.65 & 3.59× & 9.46 & 3.54× & 8.73
        & 2.32× & 5.41 & 3.50× & 8.90 \\
        & \textbf{\method{}}
        & \textbf{7.08×} & 8.62 & \textbf{6.98×} & 8.26 & \textbf{7.90×} & 9.11
        & \textbf{6.99×} & 8.61 & \textbf{7.16×} & 8.40 & \textbf{6.75×} & 7.86
        & \textbf{4.26×} & 4.58 & \textbf{6.73×} & 7.92 \\
        \midrule
        \multirow{4}{*}{Qwen3-8B} & EAGLE-3
        & 1.79× & 3.00 & 1.94× & 3.23 & 1.81× & 3.02
        & 1.89× & 3.17 & 1.57× & 2.65 & 1.69× & 2.82
        & 1.63× & 2.83 & 1.76× & 2.96 \\
        & DFlash
        & 5.62× & 6.74 & 5.15× & 6.12 & 6.08× & 7.25
        & 5.14× & 6.41 & 5.51× & 6.74 & 4.65× & 5.82
        & 2.75× & 3.16 & 4.86× & 6.03 \\
        & DDTree
        & 2.91× & 9.55 & 2.84× & 9.25 & 3.16× & 10.31
        & 2.76× & 9.50 & 2.83× & 9.63 & 2.65× & 9.02
        & 1.71× & 5.40 & 2.65× & 8.95 \\
        & \textbf{\method{}}
        & \textbf{6.64×} & 8.32 & \textbf{6.57×} & 8.11 & \textbf{7.41×} & 9.16
        & \textbf{6.47×} & 8.44 & \textbf{6.56×} & 8.76 & \textbf{6.11×} & 7.89
        & \textbf{3.92×} & 4.67 & \textbf{6.14×} & 7.91 \\
       \midrule
        \multirow{4}{*}{LLaMA-3.1-8B-Instruct} & EAGLE-3
        & 2.02× & 2.81 & 1.88× & 2.62 & 1.54× & 2.24
        & 2.04× & 2.83 & 1.91× & 2.65 & 1.87× & 2.62
        & 1.85× & 2.57 & 1.87× & 2.62 \\
        & DFlash
        & 2.65× & 3.68 & 3.06× & 4.25 & 2.64× & 3.86
        & 3.51× & 4.88 & 2.77× & 3.84 & 3.57× & 5.02
        & 2.74× & 3.82 & 2.99× & 4.19 \\
        & DDTree
        & 1.47× & 5.91 & 1.66× & 6.53 & 1.44× & 6.09
        & 1.81× & 7.26 & 1.53× & 6.02 & 1.84× & 7.44
        & 1.49× & 6.06 & 1.61× & 6.47 \\
        & \textbf{\method{}}
        & \textbf{3.50×} & 5.16 & \textbf{4.00×} & 5.78 & \textbf{3.46×} & 5.37
        & \textbf{4.46×} & 6.55 & \textbf{3.56×} & 5.27 & \textbf{4.48×} & 6.69
        & \textbf{3.61×} & 5.36 & \textbf{3.87×} & 5.74 \\
        \midrule
        \multicolumn{2}{c}{\rule[-0.6ex]{0pt}{2.8ex}NVIDIA A800-SXM4-80GB GPU} & Speedup & $\tau$ & Speedup & $\tau$ & Speedup & $\tau$
        & Speedup & $\tau$ & Speedup & $\tau$ & Speedup & $\tau$
        & Speedup & $\tau$ & Speedup & $\tau$ \\[0.6ex]
        \midrule
        \multirow{4}{*}{Qwen3-4B} & EAGLE-3
        & 1.64× & 2.79 & 1.89× & 3.22 & 1.75× & 2.99
        & 1.74× & 3.01 & 1.63× & 2.77 & 1.69× & 2.89
        & 1.70× & 2.95 & 1.72× & 2.95 \\
        & DFlash
        & 3.73× & 4.92 & 4.71× & 6.00 & 5.09× & 6.67
        & 4.74× & 6.04 & 4.90× & 6.50 & 4.42× & 5.66
        & 2.67× & 4.07 & 4.24× & 5.69 \\
        & DDTree
        & 4.23× & 9.36 & 5.87× & 9.26 & 5.66× & 10.52
        & 5.56× & 9.44 & 5.42× & 8.87 & 5.68× & 9.00
        & 3.53× & 5.79 & 5.13× & 8.89 \\
        & \textbf{\method{}}
        & \textbf{4.82×} & 8.19 & \textbf{6.62×} & 8.23 & \textbf{6.55×} & 9.41
        & \textbf{6.41×} & 8.46 & \textbf{6.67×} & 7.93 & \textbf{6.25×} & 7.97
        & \textbf{4.04×} & 5.00 & \textbf{5.91×} & 7.89 \\
        \midrule
        \multirow{4}{*}{Qwen3-8B} & EAGLE-3
        & 1.63× & 2.74 & 1.87× & 3.12 & 1.73× & 2.91
        & 1.75× & 3.05 & 1.56× & 2.57 & 1.64× & 2.74
        & 1.58× & 2.70 & 1.68× & 2.83 \\
        & DFlash
        & 3.57× & 4.73 & 4.67× & 5.98 & 4.84× & 6.40
        & 4.32× & 5.52 & 4.93× & 6.69 & 4.04× & 5.21
        & 2.47× & 3.80 & 4.03× & 5.48 \\
        & DDTree
        & 2.96× & 9.22 & 3.97× & 9.27 & 3.95× & 10.40
        & 3.65× & 9.61 & 3.99× & 9.15 & 3.59× & 8.79
        & 2.26× & 5.68 & 3.48× & 8.88 \\
        & \textbf{\method{}}
        & \textbf{4.84×} & 8.10 & \textbf{6.55×} & 8.17 & \textbf{6.59×} & 9.40
        & \textbf{6.05×} & 8.52 & \textbf{6.74×} & 8.08 & \textbf{5.66×} & 7.67
        & \textbf{3.65×} & 4.87 & \textbf{5.72×} & 7.83 \\
        \midrule
        \multirow{4}{*}{LLaMA-3.1-8B-Instruct} & EAGLE-3
        & 1.64× & 2.34 & 1.39× & 1.92 & 1.25× & 1.84
        & 1.74× & 2.44 & 1.57× & 2.22 & 1.45× & 2.07
        & 1.29× & 1.78 & 1.48× & 2.09 \\
        & DFlash
        & 2.56× & 3.65 & 3.09× & 4.27 & 2.61× & 3.83
        & 3.48× & 4.90 & 2.77× & 3.91 & 3.52× & 5.02
        & 2.75× & 3.79 & 2.97× & 4.20 \\
        & DDTree
        & 2.25× & 5.97 & 2.41× & 6.53 & 2.20× & 6.10
        & 2.74× & 7.28 & 2.27× & 6.00 & 2.67× & 7.43
        & 2.17× & 6.05 & 2.39× & 6.48 \\
        & \textbf{\method{}}
        & \textbf{3.98×} & 5.21 & \textbf{4.38×} & 5.82 & \textbf{3.76×} & 5.34
        & \textbf{5.02×} & 6.59 & \textbf{4.06×} & 5.36 & \textbf{4.91×} & 6.66
        & \textbf{3.85×} & 5.30 & \textbf{4.28×} & 5.75 \\
        \bottomrule
    \end{tabular}
    }
    }
    \caption{Speedup Ratio and Accept Length Comparison in different models, methods, datasets and GPUs. $\tau$ denotes the average acceptance length. \textit{Avg.} denotes the average result across all seven datasets. The best performing method under the same configuration is highlighted in \textbf{bold} font.}
    \label{tab:main-results}
\end{table*}
Table~\ref{tab:main-results} reports the end-to-end decoding performance of \method{}, EAGLE-3, DFlash, and DDTree across different models, datasets, and GPU platforms. Overall, \method{} achieves the best average speedup across all settings. Compared with DFlash, \method{} increases the accepted length by verifying a selected prefix tree rather than a single sequence. Compared with DDTree, \method{} preserves most of the accepted length while reducing the verification workload and achieves higher end-to-end throughput and speedup.

\paragraph{Acceptance-Cost Tradeoff}
We analyze the single-request latency of Qwen3-8B on an NVIDIA A800-SXM4-80GB GPU, averaged over seven datasets in Table~\ref{tab:datasets}. Figure~\ref{fig:Latency} shows that DFlash is constrained by single-sequence verification, which limits the accepted length and results in a verification latency of 2.32 s. DDTree achieves a longer accepted length, but its marginal independent selection ignores path-conditioned reachability, leading to a verification latency of 2.88 s. \method{} reduces the verification latency to 1.57 s through target-aware selection, while introducing only 0.083 s of additional selection cost, which accounts for 4.2\% of the total end-to-end time and is negligible relative to the overall generation time. These results show that the gains of \method{} come from more effective candidate selection.

\paragraph{Budget-Controlled Verification.}
We examine whether the gains of \method{} come from using a larger verification budget. In the main experiments, the dynamically pruned trees produced by \method{} contain approximately 64 verification nodes on average. We therefore set DDTree to the same 64-node budget and compare the accepted length on Qwen3-4B across seven datasets. Figure~\ref{fig:budget} shows that \method{} improves the average accepted length from 6.93 to 7.92, a relative gain of 14.2\%. This result shows that, by selecting a prefix-closed subtree based on target-aware path acceptance probabilities, \method{} can retain more effective accepted tokens under the same node budget.

\begin{figure}[b]
    \centering
    \includegraphics[width=\columnwidth]{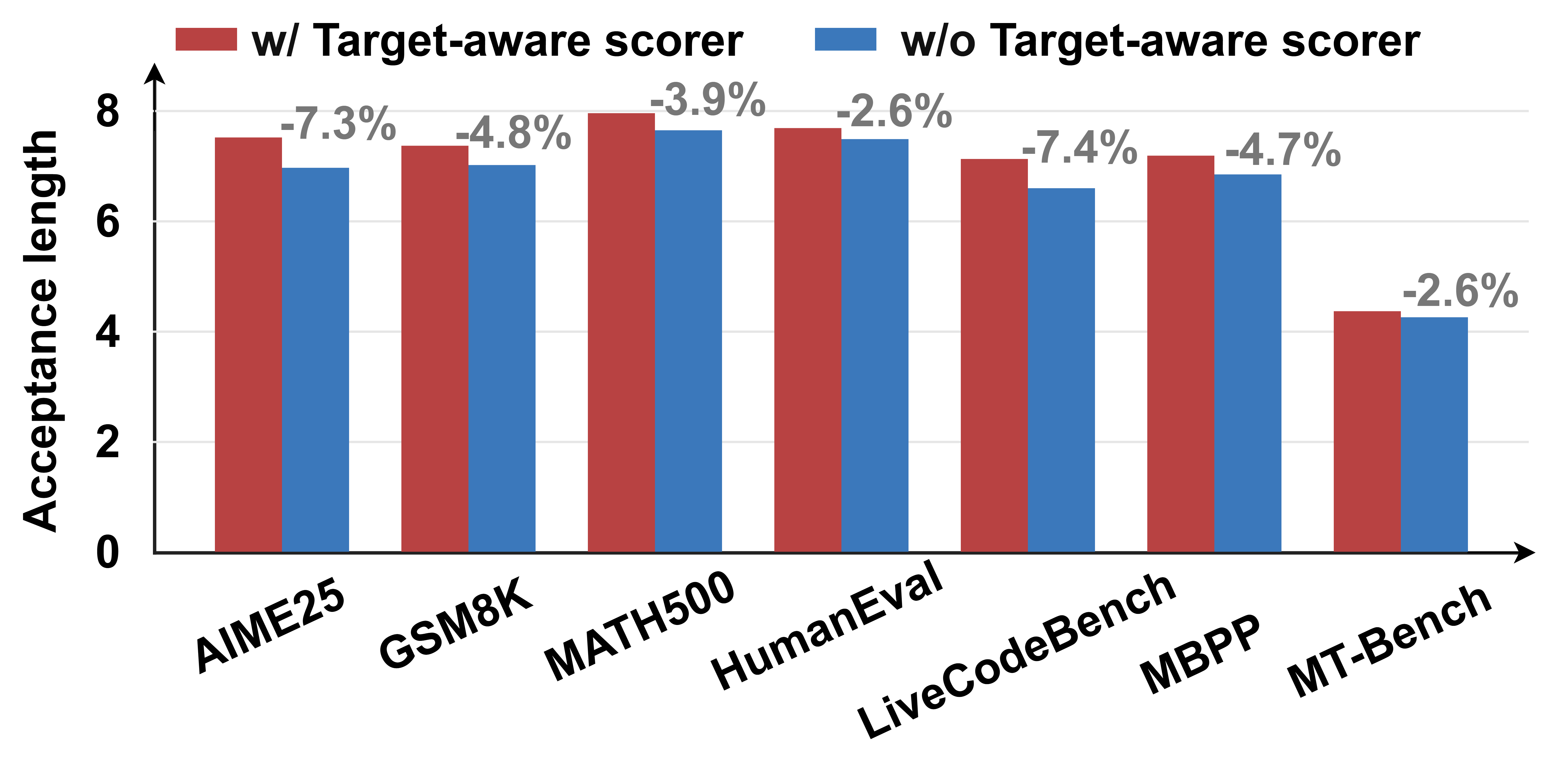}
    \caption{Effect of path-conditional scoring strategy on accepted length. }
    \label{fig:target}
\end{figure}

\begin{figure}[t]
    \centering
    \includegraphics[width=\columnwidth]{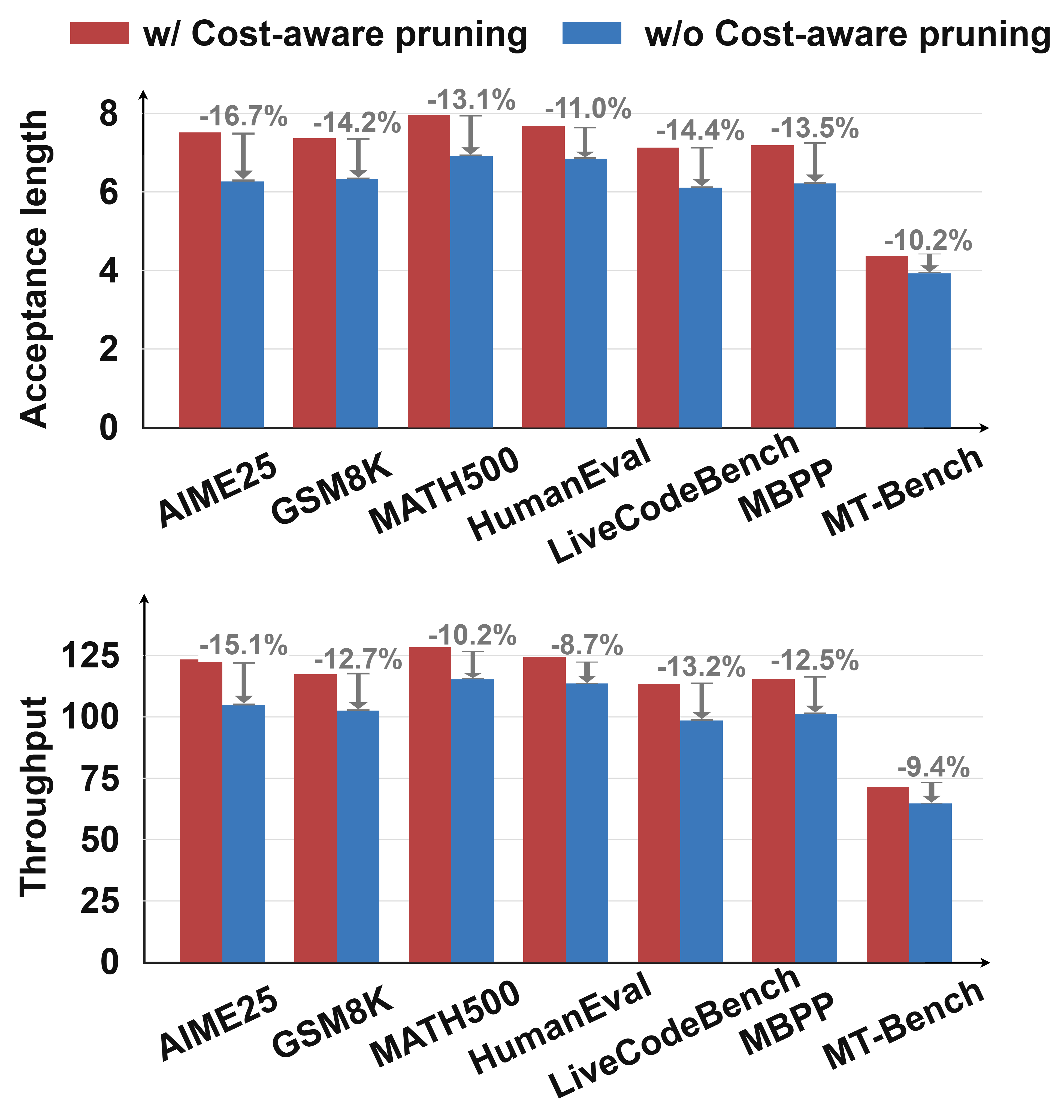}
    \caption{Effect of target-aware dynamic pruning on accepted length and throughput. Arrows denote the relative drop after disabling cost-aware pruning. }
    \label{fig:cost}
\end{figure}

\subsection{Ablation Study}

We ablate the two key components of \method{} to understand where the performance gains come from: the target-aware scorer and the cost-aware dynamic pruning strategy. These two components correspond to the two main limitations identified in Section~3: marginal draft confidence does not faithfully reflect target-model acceptance, and larger verification trees do not necessarily lead to better end-to-end throughput.

\paragraph{Effect of path-conditional scoring.}
We first evaluate whether path-conditional scoring is necessary for effective prefix-tree selection. In this ablation, we remove the scorer and fall back to selection based on the drafter's marginal confidence, while keeping the candidate pool and the remaining selection procedure unchanged. As shown in Figure~\ref{fig:target}, disabling path-conditional scoring consistently reduces the accepted length across all seven benchmarks, with relative drops ranging from 2.6\% to 7.4\% and an average drop of 4.8\%. This confirms that the improvement of \method{} is from the scorer, which helps identify prefixes that are more likely to survive target-model verification. By converting position-wise marginal confidence into path-conditioned acceptance estimates, \method{} better matches the sequential nature of speculative verification and retains more useful prefixes under the same selection process.

\paragraph{Effect of target-aware dynamic pruning.}
We then ablate the target-aware dynamic pruning strategy while keeping the path-conditional scorer unchanged. This variant still assigns target-aware scores to candidate nodes, but it no longer selects the final tree according to the expected acceptance gain per verification cost. As shown in Figure~\ref{fig:cost}, removing target-aware pruning leads to consistent degradation in both accepted length and throughput. The accepted length drops by 10.2\%--16.7\% across benchmarks, with an average drop of 13.3\%, while throughput drops by 8.7\%--15.1\%, with an average drop of 11.7\%. These results show that accurate scoring alone is not sufficient: without target-aware pruning, the selector can still spend verification budget on nodes whose expected contribution is small relative to their verification cost. In contrast, \method{} adaptively preserves a compact prefix-closed subtree according to both path reachability and verification cost, thereby improving the acceptance-cost trade-off and translating better tree selection into end-to-end throughput gains.

\section{Conclusion}

We propose \method{}, a target-aware prefix selection method for diffusion-draft speculative decoding. 
We show that the inefficiency of existing diffusion draft trees comes from path-unaware budget allocation: correct tokens are often available in marginal candidates, but large verification trees are needed to recover useful prefixes, leading to diminishing throughput returns. 
\method{} addresses this issue by estimating path-conditioned acceptance probabilities and selecting a compact cost-aware prefix tree for target-model verification. 
Extensive experiments across multiple models, datasets, and GPU platforms demonstrate that \method{} achieves up to 7.9$\times$ lossless end-to-end speedup while improving the acceptance-cost tradeoff over strong speculative decoding baselines.

\section*{Limitations}
Our evaluation focuses on deterministic greedy decoding with temperature $=0$.
In this setting, verification reduces to checking whether a draft token matches the target model's greedy choice, which makes the scorer's acceptance labels well defined.
Under sampling-based decoding, however, acceptance is no longer tied to a single greedy token: multiple candidates at the same position may be accepted depending on the sampled target output.
Extending \method{} to this setting would require recalibrating the scorer and pruning objective for sampling-specific acceptance rules.

Our current implementation is a single-request research prototype and is not yet integrated into production serving systems such as vLLM~\citep{kwon2023vllm} or SGLang~\citep{zheng2024sglang}. In batched serving, dynamic draft trees interact with continuous batching, paged attention, KV-cache management, and request scheduling, changing the cost profile of tree-structured verification.
We leave production-level integration of dynamic tree selection in multiple batches systems to future work.

Finally, \method{} operates on the candidate pool produced by the diffusion drafter and improves only the selection stage, not the draft quality itself.
When the drafter's top-$K$ marginal predictions fail to cover the target model's preferred tokens, for instance, on out-of-distribution prompts, the ceiling on acceptance length is set by the candidate pool, and better selection alone cannot compensate.
\method{} is thus complementary to efforts that improve drafter accuracy, and combining the two directions is a promising avenue for future work.

\bibliography{custom}

\clearpage

\end{document}